\documentclass[journal]{IEEEtran}

\usepackage{cite}
\usepackage{amsmath,amsfonts,amssymb}
\usepackage{algorithm}
\usepackage{algorithmicx}
\usepackage{algpseudocode}
\usepackage{array}
\usepackage{textcomp}
\usepackage{url}
\usepackage{graphicx}

\usepackage{booktabs}       
\usepackage{tabularx}       
\usepackage{threeparttable} 
\usepackage{multirow}       
\usepackage{makecell}       

\usepackage{bm}             

\usepackage[dvipsnames]{xcolor}
\usepackage[hidelinks]{hyperref}
\usepackage{orcidlink}
\usepackage{cleveref}       
\crefformat{figure}{Fig.~#2#1#3}            
\crefformat{table}{Table~#2#1#3}            
\crefformat{algorithm}{Algorithm~#2#1#3}    
\crefname{section}{Section}{Sections}       
\crefname{subsection}{Section}{Sections}    

\usepackage{placeins}       

\begin{document}
\title{Widest-Path Reachability Fields for Connectivity-Preserving Slender Structure Segmentation}

\author{
  Youcheng Zong\(^{\orcidlink{0009-0008-6795-8412}}\),~\IEEEmembership{Student~Member,~IEEE},
  Runda Jia\(^{\orcidlink{0000-0002-8586-243X}}\),
  Minxuan Hu\(^{\orcidlink{0009-0002-0059-338X}}\),
  Weilan Su\(^{\orcidlink{0009-0002-9694-7487}}\),
  \\ and Dakuo He\(^{\orcidlink{0000-0001-8303-529X}}\)
  \thanks{This work was supported by the Fundamental Research Funds for the Central Universities, China (N26GFZ006). \textit{(Corresponding author: Runda Jia.)}}
  \thanks{Youcheng Zong, Runda Jia, Minxuan Hu, and Dakuo He are with the College of Information Science and Engineering, Northeastern University, Shenyang 110004, China (e-mail: zongyc@mails.neu.edu.cn; jiarunda@ise.neu.edu.cn; humingxuan@mails.neu.edu.cn; hedakuo@ise.neu.edu.cn).}
  \thanks{Weilan Su is with the Changsha Stomatological Hospital, Changsha 410005, China (e-mail: suweilan@zsskqyy1.wecom.work).}
  \thanks{The source code is available at \href{https://github.com/mituan-ai/WPRF_open}{https://github.com/mituan-ai/WPRF\_open}.}
}

\markboth{Preprint, July 2026}
{Widest-Path Reachability Fields for Connectivity-Preserving Slender Structure Segmentation}

\maketitle

\begin{abstract}
  Segmenting slender curvilinear structures such as retinal vessels, cracks, and roads demands topological correctness, as even a single-pixel discontinuity can fragment a continuous network and invalidate downstream analysis.
  Under standard binary-mask supervision, models optimized for pixel-level overlap frequently produce topologically broken predictions.
  We trace this to a fundamental mismatch: pixel-wise losses distribute gradients uniformly, yet connectivity hinges on a sparse set of bottleneck pixels. These pixels are vastly outnumbered by thick structures and background, rendering their aggregate gradient contribution negligible. We term this phenomenon topological gradient starvation (TGS).
  To address it, we propose Widest-Path Reachability Fields (WPRF), a differentiable Max-Min reachability objective that redirects gradient flow to connectivity bottlenecks. The module is plug-and-play, backbone-agnostic, and incurs no inference overhead.
  WPRF implements a differentiable Max-Min objective via dynamic programming on a domain-restricted graph, coupled with a bottleneck-aware observation term that balances gradient contributions across varying structures. Compared to prior topology-aware losses that rely on post-hoc skeletonization or homology computation, WPRF directly optimizes end-to-end reachability via differentiable Max-Min algebra, enabling gradient flow to concentrate on connectivity bottlenecks without auxiliary structures.
  We introduce OMVIS, a new oral microvessel segmentation dataset. Experiments across nine architectures and six datasets validate the bottleneck-focused gradient routing mechanism. WPRF improves 87\% of experiments with fixed hyperparameters and achieves clDice gains of 7.2 percentage points on structurally fragile datasets.
\end{abstract}


\section{Introduction}\label{sec:1}

Segmenting slender curvilinear structures in medical imaging and remote sensing requires topological correctness~\cite{zhang2024anatomy,sun2025rsfconv,zao2024topology}.
Studies on vessel and road segmentation show that this requirement holds across different network structures~\cite{wen2025universalvessel,liu2023aerialroad}.
Work on context modeling and structure distillation in remote-sensing semantic segmentation addresses the same structure-preservation issue beyond medical images~\cite{fang2024contextremote,zhou2026semanticprompt}.
Even promptable foundation models such as SAM2 exhibit frequent topological breaks in zero-shot capillary segmentation~\cite{ravi2025sam2,su2025zero}.
A single micro-break fragments a continuous tree into disconnected pieces, invalidating downstream tasks such as bifurcation analysis or route planning.
Annotations are often union-only binary regions without reliable instance-level definitions, making connectivity and topological quality more important than simple region overlap~\cite{jing2020coarsetofine,chen2024spatialstructure}.
Even when the final output is a single semantic mask, breaks or false links are amplified by downstream processing such as skeletonization and length measurement.

Existing segmentation models are largely optimized for pixel-level overlap such as IoU or Dice~\cite{zhang2022guidedfilter,feng2021doublesimilarity,wang2023geometricboundary}; even with high overall scores, their predictions frequently exhibit breaks or spurious links~\cite{nan2024fuzzyairway}.
Investigating this high-overlap yet low-connectivity phenomenon across five public datasets, we observed that most topological errors localize in extremely sparse foreground bottlenecks~\cite{qi2023dynamic,kirchhoff2024skeleton}.
Under standard pixel-wise losses, bottleneck pixels generate gradients comparable to thick structures, yet their overall contribution is negligible due to extreme scarcity.
This reveals a fundamental mismatch: uniform pixel-wise losses allow abundant easy samples to dominate the gradient, overwhelming the few pixels critical for connectivity.

We term this mismatch \textit{Topological Gradient Starvation (TGS)}: under standard sum-based losses such as Dice and Binary Cross Entropy (BCE), connectivity-critical bottleneck pixels receive disproportionately weak gradient updates, making topological breaks more likely despite high overall pixel accuracy.
Inspired by the concept of gradient starvation in classification~\cite{li2021gradient}, we observe that standard segmentation losses such as Dice or BCE are Sum-operators that distribute gradients across all pixels.
Because connectivity-critical bottleneck pixels are extremely sparse, they are numerically overwhelmed by the background and easy samples such as thick vessels, receiving vanishingly small gradients.
Pixel-wise hard examples with low confidence are not equivalent to topologically critical points that sever connectivity. A pixel can be confidently predicted yet sit at a connectivity bottleneck.
Simple reweighting or hard-mining therefore cannot reliably repair breaks.
The network prioritizes learning simple textural statistics while starving the structural features essential for topology.

To address this, we redistribute gradients by changing the algebraic structure used to model reachability, rather than increasing network complexity or reweighting samples.
We propose Widest-Path Reachability Fields (WPRF).
We adopt an inference-consistent threshold connectivity criterion: two points are reachable if there exists a path whose edge weights all exceed a threshold.
Unlike sum or product formulations, WPRF employs Max-Min algebra to model reachability.
Its mathematical structure induces a bottleneck-dominated gradient flow: the back-propagated signal is dominated by the weakest bottleneck edge along the path, letting the network ignore already connected regions and concentrate updates near the current break.
This mechanism yields a dynamic seek-and-repair pattern: as bottlenecks progressively strengthen, the gradient focus automatically shifts to the next weakest link.

To close the training loop, we further introduce a domain-restricted support graph that blocks background shortcuts and a bottleneck-aware observation term that balances gradient contributions between thick and thin structures, keeping gradients both focused and directionally consistent.

\begin{figure}[!t]
  \centering
  \includegraphics[width=\linewidth]{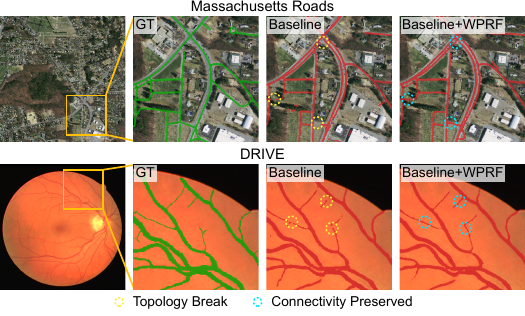}
  \caption{Qualitative comparison between a standard baseline and its +WPRF variant on Massachusetts Roads and DRIVE.}
  \label{fig:intro_motivation}
\end{figure}

Integrating WPRF into a standard Mask2Former baseline trained with the official recipe repairs typical topological breaks on both remote sensing and medical benchmarks, as shown in \cref{fig:intro_motivation}.
Experiments across nine architectures and six datasets validate this mechanism: WPRF improves clDice in 87\% of configurations with fixed hyperparameters, achieving the most substantial gains on structurally fragile datasets where bottleneck pixels are most sparse and critical, up to +7.2\,pp and +13.5\% relative improvement.
The main contributions of this paper are as follows.
\begin{itemize}
  \item
        We identify and formalize Topological Gradient Starvation (TGS): under standard pixel-wise losses, sparse connectivity-critical pixels receive insufficient gradients, causing topological breaks.
  \item
        We propose Widest-Path Reachability Fields (WPRF), which redirects gradient flow to connectivity bottlenecks through differentiable Max-Min dynamic programming on domain-restricted graphs, enabling end-to-end training without auxiliary structures.
  \item
        We validate WPRF as a plug-and-play, backbone-agnostic module with no inference overhead, improving topological connectivity of fragile structures across nine architectures and six datasets.
\end{itemize}

The remainder of this paper is organized as follows: \cref{sec:2} reviews related work, \cref{sec:3} presents our method, \cref{sec:4} reports experiments, and \cref{sec:5} concludes the paper.

\section{Related Work}\label{sec:2}
Existing methods for connectivity-preserving segmentation differ fundamentally in their underlying algebraic structures, which determine how learning signals propagate to critical bottleneck pixels.
We organize related work along three operator paradigms: sum-based, product-based, and Max-Min methods.

\subsection{Accumulative Losses: The Sum-based Paradigm}
Many segmentation methods rely on objectives governed by the sum operator. These pixel-wise objectives are typically built upon encoder-decoder architectures such as U-Net~\cite{ronneberger2015u} and modern attention or Transformer variants~\cite{tomar2023fanet,du2024swinpanet}.
Classic pixel-wise losses, such as BCE and Dice loss, compute global loss by accumulating errors across all pixels~\cite{zhang2024anatomy,zhang2022guidedfilter}. Li et al.~\cite{li2021gradient} show that sum-based objectives lead networks to prioritize easy samples.
In classification, this means neglecting rare classes; in slender structure segmentation, bottleneck pixels receive insufficient gradients.
In high-resolution images, connectivity-critical bottleneck pixels often occupy only a small fraction of the foreground, while thick regions dominate the loss and gradients.
This imbalance causes networks to learn local statistics that improve overlap metrics quickly, but leave breaks at sparse bottlenecks unrepaired~\cite{tan2020equalization,liang2022polyloss,zong2025hybridgrid}.
Methods incorporating geometric priors, such as Geodesic Distance Transform~\cite{cohen1997global} or CAPE-style connectivity-aware losses~\cite{esmaeilzadeh2025cape}, aggregate path costs additively. While they add spatial constraints, their gradient allocation still diffuses learning signals broadly.

\subsection{Decay-prone Affinities: The Product-based Paradigm}
Graph-based methods and affinity learning adopt logic based on the product operator, modeling connectivity as the product of edge probabilities.
Random Walks~\cite{huang2022capturing} and spectral clustering~\cite{grady2006random,shi2000normalized} are representative examples. Normalized-cut-style graph partitioning derives global grouping from local affinities, and TokenCut combines this with self-supervised Transformer features~\cite{shi2000normalized,wang2023tokencut}.
On large-scale 3D volumes, recent work studies efficient affinity learning and distillation to improve connectivity consistency~\cite{liu2024cross,chen2025efficient,zong2025metacontrastive}.
However, in long-path settings, multiplicative probability chains are more sensitive to local errors and may suffer from length-dependent attenuation and numerical instability~\cite{arroyo2025vanishing,cheng2024sagman}.

\subsection{Auxiliary Topological Regularizers}
Explicit topological regularizers have been explored along several directions~\cite{xu2024toposemiseg,berger2024topomulticlass,wyburd2024anatomically}. One family relies on skeleton and centerline supervision, including Skeleton Recall Loss, clCE, cbDice, and self-supervised skeleton completion~\cite{kirchhoff2024skeleton,acebes2024centerline,shi2024centerline,ren2024skeletoncompletion}.
Another family repairs breaks via explicit connectivity modeling, boundary-aware detection, or iterative tracing. Representative methods include Dynamic Snake Convolution, GLCP, NETracer, Curvi-Tracker, and crack-focused boundary-aware models~\cite{qi2023dynamic,zhou2025glcp,liu2025netracer,heng2026curvitracker,wu2025boundarycrack,qu2022deepcrackdscnn}. Other approaches exploit structure-aware representations and topology-constrained reconstruction, such as Fractal Feature Maps, HarmonySeg, and VesselSDF~\cite{huang2024fractal,huang2025harmonyseg,esposito2025vesselsdf}.
Topology-driven supervision has been extended to domain-specific and cross-domain settings, including multi-stage coronary artery extraction~\cite{qiu2025coronary}, structure-constrained weakly supervised segmentation~\cite{chen2024spatialstructure}, and topology-aware constraints under test-time adaptation~\cite{zhou2025topotta}.

clDice~\cite{shit2021cldice} extracts centerlines via skeletonization to compute overlap, while TopoLoss~\cite{clough2022topological}, topology-aware focal loss~\cite{demir2023topologyaware}, and SATLoss~\cite{wen2025topology} penalize topological changes based on persistent homology. However, skeletonization is typically non-differentiable or requires approximate gradients, and persistent homology is computationally expensive.
To reduce overhead and improve differentiability, recent works introduce alternative signals, such as morphology-inspired priors, Euler-characteristic-based objectives, and topology-preserving downsampling~\cite{wu2024deepclosing,li2025euler,chen2024topodownsample}.
A common limitation is that these methods act as morphological constraints atop pixel-wise primary objectives. Their optimization targets may not align with the threshold-based connectivity criterion used at inference, and their gradients are still aggregated through sum-based operators.
WPRF uses a Max-Min operator to directly align the training objective with threshold-based connectivity at inference.

\subsection{Bottleneck-Focused Methods: The Max-Min Paradigm}
In contrast to the above methods, WPRF adopts bottleneck-focused modeling based on Max-Min algebra for connectivity-preserving segmentation.
In graph theory, the Widest Path Problem seeks to maximize the minimum edge weight along a path~\cite{ahuja1993networkflows}. This work explores how to systematize this mathematical structure into an end-to-end trainable segmentation objective.
Unlike sum-based methods that diffuse gradients and product-based methods that suffer from attenuation, Max-Min directly encodes the bottleneck-limiting nature of connectivity: a path's strength is determined by its weakest link. This algebraic property induces a bottleneck-dominated gradient flow, where learning signals concentrate on connectivity-critical edges.
This gradient concentration is not a heuristic pixel-reweighting scheme, but an inherent consequence of the Max-Min operator's mathematical structure.

\section{Method}\label{sec:3}
We introduce the WPRF framework and its training and inference protocol.
\cref{sec:3-1} formulates the problem setup and the architectural overview,
\cref{sec:3-2} describes the support graph construction bridging pixel and graph domains,
\cref{sec:3-3} defines the widest-path reachability field and its gradient routing property,
and \cref{sec:3-4} presents the training objectives and the fixed-threshold inference protocol.

\subsection{Overall Framework}\label{sec:3-1}
Given an input image $I\in\mathbb{R}^{H_0\times W_0\times 3}$, we predict a foreground mask of slender structures from binary-mask supervision, optimizing threshold-based connectivity without instance labels or skeleton annotations.
We denote the GT union as a binary mask $U^\ast:\Omega_0\to\{0,1\}$ on the pixel grid $\Omega_0=\{1,\dots,H_0\}\times\{1,\dots,W_0\}$ and the predicted foreground probability as $U(x)=\sigma(u(x))\in(0,1)$.

Pixel-wise losses optimize local overlap but lack structural sensitivity to connectivity bottlenecks; this is the TGS problem discussed in \cref{sec:1}.
Pixel-level reweighting and hard-negative mining address numerical imbalance but not structural connectivity. Product-based formulations such as random walks suffer exponential signal decay over long paths.
We adopt Max-Min algebra, where a path's strength equals its weakest edge, concentrating gradients on the current connectivity bottleneck.

\begin{figure*}[t]
  \centering
  \includegraphics[width=\linewidth]{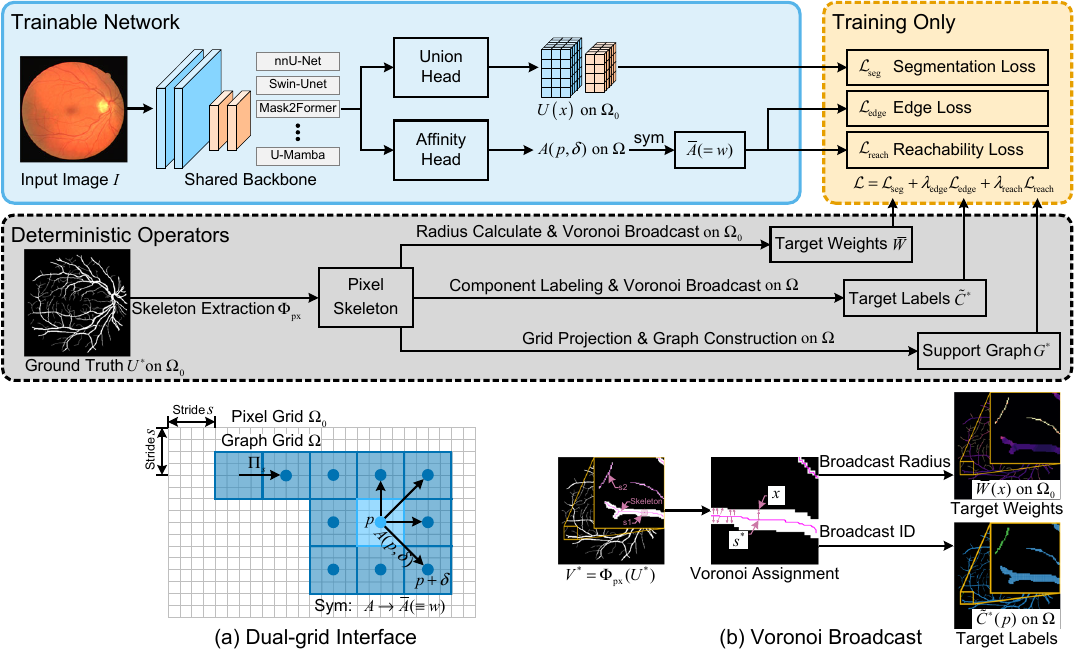}
  \caption{Architecture of the proposed WPRF framework. A shared backbone predicts union foreground probabilities $U(x)$ on the pixel grid $\Omega_0$ and directed affinities $A(p,\delta)$ on the graph grid~$\Omega$. Deterministic operators extract a skeleton-based support domain and build a domain-restricted support graph; skeleton-derived targets supervise $\mathcal{L}_{\mathrm{seg}}$ and $\mathcal{L}_{\mathrm{edge}}$, while widest-path reachability on the graph supervises $\mathcal{L}_{\mathrm{reach}}$. At inference, only the thresholded union mask is output without graph post-processing. (a)~dual-grid interface aligning $\Omega_0$ and $\Omega$; (b)~Voronoi broadcast for densifying skeleton-derived targets.}
  \label{fig:wprf_method_overview}
\end{figure*}

\cref{fig:wprf_method_overview} shows a shared backbone feeding two heads:
a union head producing pixel-level foreground probabilities $U(x)$ and
an affinity head producing directed link probabilities $A(p,\delta)$ on a coarser graph grid~$\Omega$.
This design is backbone-agnostic; \cref{sec:4} evaluates its compatibility with different backbones.
From the GT union mask, we extract a skeleton-based support domain $V^\ast\subseteq\Omega$ and build a support graph $G^\ast=(V^\ast,E^\ast)$.
Widest-path reachability on this graph provides a differentiable Max-Min training signal routing learning toward connectivity bottlenecks.
At inference, we output only the thresholded union mask $\hat U=\mathbf{1}[U>\tau_{\mathrm{fg}}]$ without graph post-processing.
\cref{alg:wprf} gives the full procedure.

\subsection{Support Graph Construction}\label{sec:3-2}
We describe how pixel-domain predictions are bridged to graph-domain structural supervision.

We define a graph grid $\Omega=\{1,\dots,H'\}\times\{1,\dots,W'\}$ with stride~$s$, where $H'=H_0/s$ and $W'=W_0/s$.
We assume $H_0$ and $W_0$ are divisible by $s$; padding is applied otherwise.
To align pixel-domain quantities to this grid, we use cell-wise max-pooling $\Pi_s$: for any scalar field $X:\Omega_0\to\mathbb{R}$,
\begin{IEEEeqnarray}{rCl}
  (\Pi_s X)(y,x) & = & \max_{0\le i<s,\, 0\le j<s} \nonumber\\
  && X\big(s(y{-}1){+}i{+}1,\, s(x{-}1){+}j{+}1\big). \label{eq:proj}
\end{IEEEeqnarray}
When $X$ is binary, $\Pi_s$ reduces to occupancy projection.
A deterministic operator $\Phi_{\text{px}}$ extracts a one-pixel-wide skeleton from the union foreground.
In practice, we apply lightweight morphological preprocessing to fill small gaps in the union foreground and use a deterministic thinning operator to extract a one-pixel-wide skeleton.
The GT support domain on the graph grid is:
\begin{IEEEeqnarray}{rCl}
  V^\ast & = & \Pi_s(\Phi_{\text{px}}(U^\ast))\subseteq\Omega. \label{eq:support_domain}
\end{IEEEeqnarray}

The union head outputs logits $u(x)\in\mathbb{R}$ on~$\Omega_0$, yielding $U(x)=\sigma(u(x))$.
The affinity head outputs directed link logits $a(u,\delta)\in\mathbb{R}$ for each node $u\in\Omega$ and offset $\delta\in\mathcal{O}$, where $\mathcal{O}\subset\mathbb{Z}^2$ is a fixed, origin-symmetric neighborhood. Tthe directed link probability is $A(u,\delta)=\sigma(a(u,\delta))\in(0,1)$.
The neighborhood $\mathcal{O}$ induces an edge set $E$ on $\Omega$, restricted to $E^\ast$ over the GT support domain:
\begin{IEEEeqnarray}{rCl}
  E & = & \{(u,v)\mid v=u+\delta,\ \delta\in\mathcal{O}\}, \nonumber\\
  E^\ast & = & \{(u,v)\in E\mid u,v\in V^\ast\}.
\end{IEEEeqnarray}
For any pair $v=u+\delta$, we symmetrize to obtain the undirected link probability $\bar A_{uv}$, which serves as the edge weight $w_{uv}$:
\begin{IEEEeqnarray}{rCl}
  \bar A_{uv} & = & \frac{A(u,\delta)+A(v,-\delta)}{2},\qquad v=u+\delta, \nonumber\\
  w_{uv} & = & \bar A_{uv}.
\end{IEEEeqnarray}

Structural reachability is defined only within a domain mask $M\subseteq\Omega$; edges touching nodes outside~$M$ are discarded:
\begin{IEEEeqnarray}{l}
  w^M_{uv}=w_{uv}\cdot\mathbf{1}[u\in M\wedge v\in M].
\end{IEEEeqnarray}
Here $\mathbf{1}[\cdot]$ is the indicator function.
During training $M=V^\ast$; during inference $M=\hat V=\Pi_s(\Phi_{\text{px}}(\hat U))$.
This restriction prevents paths from detouring through background, which would create spurious shortcuts between topologically distinct structures.

\subsection{Widest-Path Reachability Fields}\label{sec:3-3}
On the domain-restricted support graph, connectivity is determined by the weakest edge along a path. We adopt the classic widest-path max--min connectivity formulation~\cite{udupa1996fuzzyconnectedness,ahuja1993networkflows} as a differentiable training signal.
For a path $\pi=(v_0{=}u,v_1,\dots,v_t{=}v)$ within~$M$, define its path strength as the bottleneck weight:
\begin{IEEEeqnarray}{rCl}
  S(\pi) & = & \min_{0\le i<t}\,w^{M}_{v_iv_{i+1}}. \label{eq:path_strength}
\end{IEEEeqnarray}
The $k$-step widest-path reachability field (WPRF) is
\begin{IEEEeqnarray}{l}
  r^{(k)}_{\mathrm{mm}}(u,v)=\max_{1\le t\le k}\;\max_{\pi:\,|\pi|=t}S(\pi)\;\in[0,1]. \label{eq:wprf}
\end{IEEEeqnarray}
Here $|\pi|$ denotes the number of edges in the path.
This definition matches the threshold connectivity criterion used at inference: for any threshold~$\tau$, two nodes are connected by a path of length~${\le}k$ in the thresholded subgraph $\{(u,v)\mid w^M_{uv}>\tau\}$ if and only if $r^{(k)}_{\mathrm{mm}}(u,v)>\tau$.
Training against this target aligns the learned affinities with the inference-time criterion; \cref{sec:4} verifies this via ablations.

We compute $r^{(k)}_{\mathrm{mm}}$ by $k$-step dynamic programming in the Max-Min semiring. For source~$u$, we initialize $r_0$ and iterate:
\begin{IEEEeqnarray}{rCl}
  r_0(v) & = &
  \begin{cases}
    1, & v=u,              \\
    0, & \text{otherwise},
  \end{cases}
  \label{eq:dp_init}
\end{IEEEeqnarray}
\begin{IEEEeqnarray}{l}
  r_{t+1}(v)=\max\!\Big(r_t(v),\;\max_{p:(p,v)\in E}\min\big(r_t(p),\,w^M_{pv}\big)\Big), \label{eq:dp}
\end{IEEEeqnarray}
for $t=0,\dots,k{-}1$, and set $r^{(k)}_{\mathrm{mm}}(u,v)=r_k(v)$.
The process uses only $\min$/$\max$ operations, which are almost-everywhere differentiable; at ties, automatic differentiation returns a valid subgradient.
When the optimal path and its bottleneck edge are unique, gradients propagate only through that bottleneck. When multiple paths tie, the implementation returns one valid subgradient.

\begin{figure}[t]
  \centering
  \includegraphics[width=\linewidth]{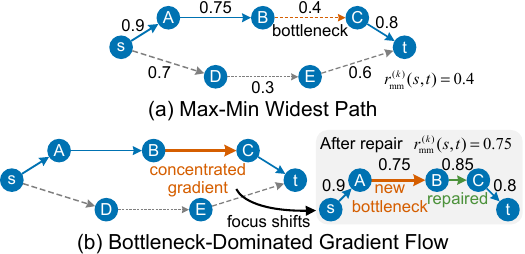}
  \caption{Bottleneck-dominated gradient routing induced by Max-Min reachability. (a)~On a toy graph with edge weights $w\in[0,1]$, the $k$-step widest-path reachability $r^{(k)}_{\mathrm{mm}}(s,t)$ equals the minimum edge weight along the widest path (the bottleneck). (b)~Back-propagation through $\max$--$\min$ concentrates gradients on the current bottleneck edge; once strengthened, the maximizer switches to the next weakest link, producing a ``seek-and-repair'' pattern.}
  \label{fig:wprf_propagation}
\end{figure}

\textit{Bottleneck-dominated gradient routing:} Since $\min$ retains only the bottleneck edge per candidate path, back-propagated gradients concentrate on the weakest edge limiting reachability.
Once that bottleneck is strengthened, the maximizer may switch to a new path and bottleneck, shifting learning to the next weak link, see \cref{fig:wprf_propagation}.
This routing effect is an inherent property of the Max-Min algebra, not an external reweighting trick.
\cref{sec:4} provides mechanistic analyses and ablations confirming this behavior.
The max--min connectivity formulation itself is classical~\cite{udupa1996fuzzyconnectedness,ahuja1993networkflows}. Our contribution is making it a differentiable training objective via $k$-step dynamic programming (DP) over learned affinities, coupled with a domain-restricted support graph and a bottleneck-aware observation term closing the training loop for segmentation.

\subsection{Training and Inference}\label{sec:3-4}
We optimize three objectives jointly and use a fixed-threshold inference protocol.

We sample point pairs on the GT support graph $G^\ast=(V^\ast,E^\ast)$ to supervise structural reachability.
For each image, we uniformly sample $N_s$ source nodes from $V^\ast$ and pair each source with targets satisfying distance criteria in $P_k$ and $N_k$.
Let $\mathrm{CC}^\ast(\cdot)$ denote connected-component IDs and $\mathrm{dist}_{G^\ast}(\cdot,\cdot)$ the shortest-path distance on $G^\ast$.
For brevity, we write $\mathrm{dist}(u,v)$ for $\mathrm{dist}_{G^\ast}(u,v)$.
For each scale $k\in\mathcal{K}$, positive pairs are same-component nodes with $\mathrm{dist}\in[\lceil k/2\rceil,k]$;
the lower bound $\lceil k/2\rceil$ excludes trivially reachable pairs, focusing supervision on near-disconnection bottlenecks at scale~$k$.
Negative pairs are inter-component pairs together with intra-component pairs with $\mathrm{dist}>k$; the latter penalize spatially close nodes that should not be connected within $k$ steps.
We denote the positive and negative pair sets at scale~$k$ by $P_k$ and $N_k$:
\begin{IEEEeqnarray}{rCl}
  P_k & = & \{(u,v)\in (V^\ast)^2 \mid \mathrm{CC}^\ast(u)=\mathrm{CC}^\ast(v), \nonumber\\
  && \lceil k/2\rceil \le \mathrm{dist}(u,v)\le k\}, \nonumber\\
  N_k & = & \{(u,v)\in (V^\ast)^2 \mid \mathrm{CC}^\ast(u)\ne\mathrm{CC}^\ast(v)\} \nonumber\\
  && {}\cup \{(u,v)\in (V^\ast)^2 \mid \mathrm{CC}^\ast(u)=\mathrm{CC}^\ast(v), \nonumber\\
  && \mathrm{dist}(u,v)>k\}.
\end{IEEEeqnarray}
We assign labels $y=1/0$ to positive/negative pairs to supervise $r^{(k)}_{\mathrm{mm}}(u,v)$.
Equivalently, the pair label can be written as an indicator function:
\begin{IEEEeqnarray}{rCl}
  y(u,v) & = & \mathbf{1}[(u,v)\in P_k]. \label{eq:pair_label}
\end{IEEEeqnarray}
Intra-component pairs with $\mathrm{dist}(u,v)<\lceil k/2\rceil$ are excluded from supervision at scale~$k$ as they are trivially reachable.

Segmentation loss $\mathcal{L}_{\mathrm{seg}}$.
Although $\mathcal{L}_{\mathrm{reach}}$ routes structural gradients to connectivity bottlenecks, thick regions can still dominate the pixel-level term, leaving thin bottlenecks under-supervised.
We counter this with a Bottleneck-Aware Balanced Hard-Negative BCE.
Let $\ell(x)=\mathrm{BCE}(U(x),U^\ast(x))$ and define positive/negative pixel sets $\Omega^+=\{x\mid U^\ast(x)=1\}$, $\Omega^-=\{x\mid U^\ast(x)=0\}$.
On~$\Omega^+$, we construct a bottleneck-aware weight based on the dense skeleton-radius field $r_{\mathrm{dense}}(x)$ and normalize it:
\begin{IEEEeqnarray}{rCl}
  \tilde W(x) & = & \frac{1}{r_{\mathrm{dense}}(x)+\epsilon},\qquad x\in\Omega^+, \nonumber\\
  \bar W(x) & = & \frac{\tilde W(x)}{\mathbb{E}_{x\in\Omega^+}[\tilde W(x)]}. \label{eq:weight_def}
\end{IEEEeqnarray}
This normalization ensures $\mathbb{E}_{x\in\Omega^+}[\bar W(x)]=1$:
\begin{IEEEeqnarray}{rCl}
  \mathbb{E}_{x\in\Omega^+}[\bar W(x)] & = & 1. \label{eq:weight_norm}
\end{IEEEeqnarray}
Thinner structures receive higher per-pixel weight, ensuring bottleneck pixels are not overwhelmed by thick regions in the segmentation loss.
We use Voronoi nearest-skeleton assignment (VNA) to avoid target mixing near component boundaries and keep the procedure deterministic. This assignment is reused wherever skeleton attributes must be broadcast to dense pixels.
We compute a Euclidean distance transform of the union foreground in the same preprocessed domain as~$\Phi_{\text{px}}$, sample radii at skeleton points, and broadcast them to~$\Omega^+$ via Voronoi nearest-skeleton assignment.

To ensure determinism, ties between nearest skeleton points are resolved by a fixed rule.

Here $\epsilon>0$ is a small constant for numerical stability.
From~$\Omega^-$, we rank negatives by per-pixel loss $\ell(x)$ and select the top $|\Omega^+|$ as $\Omega^-_{\mathrm{hard}}$:
We formalize the ``top-$|\Omega^+|$'' selection as a subset maximization problem:
\begin{IEEEeqnarray}{rCl}
  \Omega^-_{\mathrm{hard}} & = & \operatorname*{arg\,max}_{S\subseteq\Omega^-,\ |S|=|\Omega^+|}\ \sum_{x\in S}\ell(x). \label{eq:hard_negatives}
\end{IEEEeqnarray}
\begin{IEEEeqnarray}{l}
  \mathcal{L}_{\mathrm{seg}}=\frac{\sum_{x\in\Omega^+}\bar W(x)\,\ell(x)+\sum_{x\in\Omega^-_{\mathrm{hard}}}\ell(x)}{|\Omega^+|+|\Omega^-_{\mathrm{hard}}|}. \label{eq:lseg}
\end{IEEEeqnarray}
This term handles speckles and scale imbalance; its contribution relative to the structural terms $\mathcal{L}_{\mathrm{edge}}$ and $\mathcal{L}_{\mathrm{reach}}$ is isolated in \cref{sec:4} ablations.

Local edge loss $\mathcal{L}_{\mathrm{edge}}$.
We supervise the symmetrized link probability $\bar A_{uv}$ directly to be high within the same component and low across different components.
GT skeleton component IDs are broadcast to the union occupancy domain $U^\ast_\Omega=\Pi_s(U^\ast)$ via VNA, producing a dense component field~$\tilde C^\ast$. The assignment operates only inside~$U^\ast_\Omega$ and preserves original IDs on~$V^\ast$.
Ties are resolved by the same fixed rule to ensure determinism.
An edge $(u,v{=}u{+}\delta)$ is positive if $\tilde C^\ast(u)=\tilde C^\ast(v)>0$ and negative otherwise.
We denote the positive and negative edge sets $E^+$ and $E^-$.
The positive and negative edge sets can be written as:
\begin{IEEEeqnarray}{rCl}
  E^+ & = & \{(u,v)\in E\mid \tilde C^\ast(u)=\tilde C^\ast(v)>0\}, \nonumber\\
  E^- & = & E\setminus E^+.
\end{IEEEeqnarray}
Supervising only skeleton edges leaves most foreground links unlabeled during training, creating a gap with inference where the full predicted mask is used; broadcasting to the union domain closes this gap.
The loss averages BCE over positive and negative edges with equal weight:
\begin{IEEEeqnarray}{l}
  \mathcal{L}_{\mathrm{edge}}=\tfrac{1}{2}\Big(\mathbb{E}_{E^+}\!\big[\mathrm{BCE}(\bar A_{uv},1)\big]+\mathbb{E}_{E^-}\!\big[\mathrm{BCE}(\bar A_{uv},0)\big]\Big). \IEEEeqnarraynumspace \label{eq:ledge}
\end{IEEEeqnarray}

Structural reachability loss $\mathcal{L}_{\mathrm{reach}}$.
For each scale $k\in\mathcal{K}$, we compute $r^{(k)}_{\mathrm{mm}}(u,v)$ on $M^\ast=V^\ast$ for sampled pairs and apply $\mathrm{BCE}(r^{(k)}_{\mathrm{mm}}(u,v),y)$ against the pair label~$y$, penalizing bottleneck edges causing disconnection.
We denote the per-scale loss by $\mathcal{L}^{(k)}_{\mathrm{reach}}$, defined as the mean BCE over sampled pairs at scale~$k$.
The multi-scale loss is
\begin{IEEEeqnarray}{rCl}
  \mathcal{L}_{\mathrm{reach}} = \sum_{k\in\mathcal{K}}\gamma_k\,\mathcal{L}^{(k)}_{\mathrm{reach}},\,\sum_{k\in\mathcal{K}}\gamma_k = 1.
\end{IEEEeqnarray}

Overall objective.
\begin{IEEEeqnarray}{l}
  \mathcal{L}=\mathcal{L}_{\mathrm{seg}}+\lambda_{\mathrm{edge}}\,\mathcal{L}_{\mathrm{edge}}+\lambda_{\mathrm{reach}}\,\mathcal{L}_{\mathrm{reach}}. \label{eq:total}
\end{IEEEeqnarray}
Here $\lambda_{\mathrm{edge}}$ and $\lambda_{\mathrm{reach}}$ are loss weights.

At test time, we binarize $U$ as $\hat U=\mathbf{1}[U>\tau_{\mathrm{fg}}]$; this is the primary output.
For optional connectivity diagnostics, we extract $\hat V=\Pi_s(\Phi_{\text{px}}(\hat U))$, threshold edges at $\bar A_{uv}>\tau_{\mathrm{link}}$ to get $\hat G_\tau$, and compute connected-component IDs.
The thresholded edge set is:
\begin{IEEEeqnarray}{l}
  \hat E_\tau=\{(u,v)\mid u,v\in\hat V,\;\bar A_{uv}>\tau_{\mathrm{link}}\}. \label{eq:diag_edges}
\end{IEEEeqnarray}
This diagnostic is for analysis only and does not alter the primary output.

Training overhead comes from Max-Min propagation. With $N_s$ sources per image across all scales, the time complexity is $O(\sum_{k\in\mathcal{K}} N_s\cdot k\cdot|\mathcal{O}|\cdot|M|)$, where $N_s$ denotes the number of sources sampled per scale; the deterministic operators $\Pi_s$ and $\Phi_{\text{px}}$ need no back-propagation. The DP propagation is batched over the source dimension, and each step updates all sources and spatial positions in a single tensorized GPU kernel.
This accounts only for the WPRF propagation overhead; backbone forward/backward costs are reported separately.
At inference, WPRF adds no graph propagation; the output is the thresholded union mask.
Optional connectivity diagnostics cost $O(|\hat V|\cdot|\mathcal{O}|)$.

\textit{Scope and extensibility:} WPRF's structural supervision is decoupled from the backbone and applies to any segmentation network producing dense foreground probabilities.
In weakly- or semi-supervised settings, any binary union mask (from annotations or pseudo-labels) can be used to build the support graph and apply the structural terms.
Extension to 3D volumes requires replacing the 2D neighborhood~$\mathcal{O}$ with a 3D one; multi-class foreground can be handled by building a separate support graph per class.
For instance-level tasks, one can build a per-instance graph and propagate reachability per instance, then aggregate the structural supervision to the corresponding pixel predictions.

\begin{algorithm}[t]
  \caption{WPRF Training and Inference}\label{alg:wprf}
  \begin{algorithmic}[1]
    \Require Image $I$, GT mask $U^\ast$, network $F_\theta$
    \Ensure Predicted mask $\hat U$
    \Statex Training:
    \State $(u_{\mathrm{px}},\,a_{\mathrm{gr}})\gets F_\theta(I)$ \Comment{union + affinity logits}
    \State $U(x)\gets\sigma(u_{\mathrm{px}}(x))$;\quad $A(p,\delta)\gets\sigma(a_{\mathrm{gr}}(p,\delta))$
    \State Symmetrize $A$ to $\bar A_{uv}=\frac{A(u,\delta)+A(v,-\delta)}{2}$ for $v=u+\delta$
    \Statex Deterministic preprocessing (no gradient):
    \State $V^\ast\gets\Pi_s\!\big(\Phi_{\text{px}}(U^\ast)\big)$;\quad build $G^\ast=(V^\ast,E^\ast)$
    \State Broadcast skeleton radii $\to\bar W$;\quad component IDs $\to\tilde C^\ast$
    \State Sample multi-scale pairs $\{(u_i,v_i,y_i)\}$ on $G^\ast$
    \Statex Compute losses:
    \State $\mathcal{L}_{\mathrm{seg}}\gets$ Eq.\,\eqref{eq:lseg} with $\bar W$ and hard negatives
    \State $\mathcal{L}_{\mathrm{edge}}\gets$ Eq.\,\eqref{eq:ledge} with $\tilde C^\ast$ labels
    \For{each scale $k\in\mathcal{K}$}
    \State Propagate $r^{(k)}_{\mathrm{mm}}$ via Eq.\,\eqref{eq:dp} on $M^\ast=V^\ast$
    \EndFor
    \State $\mathcal{L}_{\mathrm{reach}}\gets\sum_k\gamma_k\,\mathrm{BCE}\!\big(r^{(k)}_{\mathrm{mm}}(u_i,v_i),y_i\big)$ over sampled pairs
    \State $\mathcal{L}\gets$ Eq.\,\eqref{eq:total};\quad update $\theta$ via $\nabla_\theta\mathcal{L}$
    \Statex
    \Statex Inference:
    \State $\hat U\gets\mathbf{1}[U>\tau_{\mathrm{fg}}]$ \Comment{primary output}
    \Statex Optional connectivity diagnostics:
    \State $\hat V\gets\Pi_s\!\big(\Phi_{\text{px}}(\hat U)\big)$
    \State $\hat E_\tau\gets\{(u,v)\mid u,v\in\hat V,\;\bar A_{uv}>\tau_{\mathrm{link}}\}$
    \State Component IDs $\gets\mathrm{CC}(\hat V,\hat E_\tau)$
  \end{algorithmic}
\end{algorithm}

\section{Experiments}\label{sec:4}
We evaluate WPRF on five public datasets and one newly released dataset. \cref{sec:4-1,sec:4-2} describe experimental settings and datasets, \cref{sec:4-3,sec:4-4,sec:4-5,sec:4-6} report main results, qualitative results, ablation studies, and analysis.

\subsection{Experimental Settings}\label{sec:4-1}

All methods are trained and evaluated in the same codebase with identical preprocessing, training recipe, and fixed hyperparameters. No per-dataset tuning is applied. \cref{tab:impl_hparams} lists all values.
Experiments run on three identical servers, each with an Intel Xeon 8470Q CPU, 90\,GB RAM, and a single NVIDIA RTX 5090 GPU (32GB). The implementation uses PyTorch 2.8.0 with CUDA 12.8. Main results report the mean over 9 runs (three repeats per server).
Each method is trained for a fixed step count, and evaluated at the final checkpoint without validation-based selection. No post-processing or test-time augmentation is applied at inference.
For each method--dataset pair, the batch size is set to the largest value that fits in GPU memory and kept identical between baseline and +WPRF.
We report two complementary metrics: the Dice coefficient $2|P\cap G|/(|P|+|G|)$ for pixel-level overlap and clDice~\cite{shit2021cldice} for skeleton-level connectivity.

\begin{table}[t]
  \centering
  \caption{Implementation details and hyperparameters. All settings are fixed across datasets without per-dataset tuning.}
  \label{tab:impl_hparams}
  {\setlength{\tabcolsep}{3pt}
    \begin{tabularx}{\linewidth}{l >{\raggedright\arraybackslash}X}
      \toprule
      Setting                                                          & Value                                               \\
      \midrule
      \multicolumn{2}{l}{\textit{Optimization \& Inference}}                                                                 \\
      Optimizer                                                        & AdamW, lr=$1{\times}10^{-4}$, wd=$1{\times}10^{-2}$ \\
      Steps                                                            & 30000                                               \\
      Augmentation                                                     & None                                                \\
      Runs                                                             & 9                                                   \\
      Inference thresholds $\tau_{\mathrm{fg}}, \tau_{\mathrm{link}}$  & 0.5, 0.5                                            \\
      \midrule
      \multicolumn{2}{l}{\textit{Graph Construction}}                                                                        \\
      Graph stride $s$                                                 & 4                                                   \\
      Neighborhood $\mathcal{O}$                                       & 8-neighborhood                                      \\
      Support-domain $\Phi_{\text{px}}$                                & closing ($3{\times}3$) + Zhang--Suen thinning       \\
      VNA tie-breaking                                                 & row-major min                                       \\
      Stability constant $\epsilon$                                    & $10^{-6}$                                           \\
      \midrule
      \multicolumn{2}{l}{\textit{Reachability Supervision}}                                                                  \\
      Multi-scale steps $\mathcal{K}$                                  & $\{1,2,4,8,16\}$, $\gamma_k{=}1/|\mathcal{K}|$      \\
      Sources per image $N_s$                                          & 32 per scale                                        \\
      Loss weights $\lambda_{\mathrm{edge}}, \lambda_{\mathrm{reach}}$ & 1.0, 1.0                                            \\
      \bottomrule
    \end{tabularx}
  }
\end{table}

We compare nine segmentation methods spanning CNN, Transformer, hybrid, and state-space architectures. All baselines are retrained in our codebase with the same recipe. The +WPRF variants add only a lightweight affinity head and the structural losses $\mathcal{L}_{\mathrm{edge}}+\mathcal{L}_{\mathrm{reach}}$ on top of the baseline; the backbone remains unchanged.

\subsection{Datasets}\label{sec:4-2}

Experiments cover six datasets spanning retinal vessels, OCT vasculature, cracks, roads, and oral capillaries. \cref{tab:dataset_summary} summarizes basic statistics.

\begin{table}[t]
  \centering
  \caption{Summary of the six evaluation datasets. Splits are denoted as train/val/test. Input size represents the resolution used in all experiments.}
  \label{tab:dataset_summary}
  {\setlength{\tabcolsep}{3pt}
    \begin{tabularx}{\linewidth}{*{2}{>{\raggedright\arraybackslash}X} c c c}
      \toprule
      Dataset                                  & Domain             & Split      & Original size    & Input size       \\
      \midrule
      DRIVE~\cite{staal2004ridge}              & Retinal vessels    & 20/--/20   & 584$\times$565   & 512$\times$512   \\
      OCTA-500 (3\,mm)~\cite{li2024octa500}    & OCTA vessels       & 140/10/50  & 304$\times$304   & 320$\times$320   \\
      OCTA-500 (6\,mm)~\cite{li2024octa500}    & OCTA vessels       & 240/10/50  & 400$\times$400   & 416$\times$416   \\
      DeepCrack~\cite{zou2019deepcrack}        & Pavement cracks    & 300/--/237 & 384$\times$544   & 384$\times$544   \\
      Massachusetts Roads~\cite{mnih2010roads} & Aerial roads       & 1108/14/49 & 1500$\times$1500 & 1024$\times$1024 \\
      OMVIS (ours)                             & Dermoscopy vessels & 149/21/42  & 1080$\times$1440 & 1088$\times$1440 \\
      \bottomrule
    \end{tabularx}
  }
\end{table}

OMVIS is an oral microvessel segmentation dataset introduced in this paper. Data collection was approved by the Medical Ethics Committee of Changsha Stomatological Hospital, Hunan Province, China (Approval No.~(2025)-Research-(008)); all participants provided written informed consent. The dataset is publicly available at \url{https://ieee-dataport.org/documents/omvis-oral-microvessel-instance-segmentation-dataset}.

The dataset comprises 212 polarized dermoscopy images (20$\times$ and 50$\times$ magnification) from 106 participants aged 18--62, recruited between December 2018 and April 2019, covering healthy controls, OSF patients at different clinical stages, and other oral mucosal lesions.

Three attending oral physicians, each with over five years of experience, independently annotated capillaries under a double-blind protocol followed by cross-validation. Targets include visible loop-like, punctate, linear, and tree-like microvessels; collecting vessels, artifacts, and blurred regions are excluded. Inter-observer Dice agreement is 0.875. The data are split at the patient level to prevent leakage.

\subsection{Main Results}\label{sec:4-3}

We compare baseline and +WPRF variants of all nine methods on all six datasets. \cref{tab:sota_drive,tab:sota_octa3,tab:sota_octa6,tab:sota_deepcrack,tab:sota_massroads,tab:sota_omvis} report results. Params/FLOPs in the tables count network forward complexity at the dataset-specific input resolution, including the lightweight affinity head in +WPRF; training-time Max-Min DP overhead is reported in \cref{tab:wprf_overhead}.

\begin{table}[t]
  \centering
  \caption{Quantitative comparison on DRIVE. Params and FLOPs are computed at dataset input resolution and include the lightweight affinity head for +WPRF rows.}
  \label{tab:sota_drive}
  {\setlength{\tabcolsep}{3pt}\scriptsize
    \begin{tabularx}{\linewidth}{l l *{4}{>{\raggedleft\arraybackslash}X}}
      \toprule
      Method                                  & Backbone & Dice $\uparrow$ & clDice $\uparrow$ & Params (M) $\downarrow$ & FLOPs (G) $\downarrow$ \\
      \midrule
      UNet~\cite{ronneberger2015u}            & resnet18 & 0.768           & 0.782             & 14.3                    & 172.7                  \\
      UNet + WPRF                             & resnet18 & 0.774           & 0.793             & 14.3                    & 174.0                  \\
      \midrule
      DeepLabv3+~\cite{chen2018deeplabv3plus} & resnet18 & 0.760           & 0.777             & 12.3                    & 92.9                   \\
      DeepLabv3+ + WPRF                       & resnet18 & 0.761           & 0.779             & 12.3                    & 97.7                   \\
      \midrule
      SegFormer~\cite{xie2021segformer}       & MiT-B1   & 0.733           & 0.735             & 13.7                    & 172.5                  \\
      SegFormer + WPRF                        & MiT-B1   & 0.757           & 0.762             & 13.7                    & 191.8                  \\
      \midrule
      Mask2Former~\cite{cheng2022mask2former} & Swin-T   & 0.789           & 0.805             & 29.8                    & 290.6                  \\
      Mask2Former + WPRF                      & Swin-T   & 0.796           & 0.831             & 29.9                    & 300.2                  \\
      \midrule
      TransUNet~\cite{chen2024transunet}      & resnet18 & 0.768           & 0.790             & 14.9                    & 259.4                  \\
      TransUNet + WPRF                        & resnet18 & 0.773           & 0.793             & 14.9                    & 264.2                  \\
      \midrule
      nnUNet~\cite{isensee2021nnunet}         & C64      & 0.788           & 0.783             & 15.9                    & 2272.4                 \\
      nnUNet + WPRF                           & C64      & 0.789           & 0.785             & 15.9                    & 2277.2                 \\
      \midrule
      CS-Net~\cite{mou2019csnet}              & C64      & 0.770           & 0.767             & 7.7                     & 1124.9                 \\
      CS-Net + WPRF                           & C64      & 0.771           & 0.770             & 7.8                     & 1129.8                 \\
      \midrule
      Swin-UNet~\cite{cao2023swinunet}        & Swin-T   & 0.784           & 0.800             & 28.0                    & 372.3                  \\
      Swin-UNet + WPRF                        & Swin-T   & 0.788           & 0.803             & 28.0                    & 377.1                  \\
      \midrule
      U-Mamba~\cite{ma2024umamba}             & C64      & 0.758           & 0.741             & 5.5                     & 824.3                  \\
      U-Mamba + WPRF                          & C64      & 0.763           & 0.748             & 5.5                     & 829.2                  \\
      \bottomrule
    \end{tabularx}
  }
\end{table}

\begin{table}[t]
  \centering
  \caption{Quantitative comparison on OCTA-500 (3\,mm).}
  \label{tab:sota_octa3}
  {\setlength{\tabcolsep}{3pt}\scriptsize
    \begin{tabularx}{\linewidth}{l l *{4}{>{\raggedleft\arraybackslash}X}}
      \toprule
      Method                                  & Backbone & Dice $\uparrow$ & clDice $\uparrow$ & Params (M) $\downarrow$ & FLOPs (G) $\downarrow$ \\
      \midrule
      UNet~\cite{ronneberger2015u}            & resnet18 & 0.814           & 0.674             & 14.3                    & 16.9                   \\
      UNet + WPRF                             & resnet18 & 0.812           & 0.679             & 14.3                    & 17.0                   \\
      \midrule
      DeepLabv3+~\cite{chen2018deeplabv3plus} & resnet18 & 0.578           & 0.599             & 12.3                    & 9.1                    \\
      DeepLabv3+ + WPRF                       & resnet18 & 0.587           & 0.613             & 12.3                    & 9.5                    \\
      \midrule
      SegFormer~\cite{xie2021segformer}       & MiT-B1   & 0.585           & 0.602             & 13.7                    & 11.0                   \\
      SegFormer + WPRF                        & MiT-B1   & 0.617           & 0.640             & 13.7                    & 12.9                   \\
      \midrule
      Mask2Former~\cite{cheng2022mask2former} & Swin-T   & 0.603           & 0.624             & 29.8                    & 29.0                   \\
      Mask2Former + WPRF                      & Swin-T   & 0.640           & 0.669             & 29.9                    & 30.0                   \\
      \midrule
      TransUNet~\cite{chen2024transunet}      & resnet18 & 0.691           & 0.656             & 14.9                    & 25.0                   \\
      TransUNet + WPRF                        & resnet18 & 0.694           & 0.668             & 14.9                    & 25.4                   \\
      \midrule
      nnUNet~\cite{isensee2021nnunet}         & C64      & 0.813           & 0.667             & 15.9                    & 221.9                  \\
      nnUNet + WPRF                           & C64      & 0.800           & 0.664             & 15.9                    & 222.4                  \\
      \midrule
      CS-Net~\cite{mou2019csnet}              & C64      & 0.796           & 0.661             & 7.7                     & 109.9                  \\
      CS-Net + WPRF                           & C64      & 0.788           & 0.657             & 7.8                     & 110.3                  \\
      \midrule
      Swin-UNet~\cite{cao2023swinunet}        & Swin-T   & 0.828           & 0.702             & 28.0                    & 36.9                   \\
      Swin-UNet + WPRF                        & Swin-T   & 0.798           & 0.705             & 28.0                    & 37.4                   \\
      \midrule
      U-Mamba~\cite{ma2024umamba}             & C64      & 0.828           & 0.669             & 5.5                     & 80.5                   \\
      U-Mamba + WPRF                          & C64      & 0.831           & 0.671             & 5.5                     & 81.0                   \\
      \bottomrule
    \end{tabularx}
  }
\end{table}

\begin{table}[t]
  \centering
  \caption{Quantitative comparison on OCTA-500 (6\,mm).}
  \label{tab:sota_octa6}
  {\setlength{\tabcolsep}{3pt}\scriptsize
    \begin{tabularx}{\linewidth}{l l *{4}{>{\raggedleft\arraybackslash}X}}
      \toprule
      Method                                  & Backbone & Dice $\uparrow$ & clDice $\uparrow$ & Params (M) $\downarrow$ & FLOPs (G) $\downarrow$ \\
      \midrule
      UNet~\cite{ronneberger2015u}            & resnet18 & 0.794           & 0.638             & 14.3                    & 28.5                   \\
      UNet + WPRF                             & resnet18 & 0.789           & 0.641             & 14.3                    & 28.7                   \\
      \midrule
      DeepLabv3+~\cite{chen2018deeplabv3plus} & resnet18 & 0.553           & 0.564             & 12.3                    & 15.3                   \\
      DeepLabv3+ + WPRF                       & resnet18 & 0.554           & 0.571             & 12.3                    & 16.1                   \\
      \midrule
      SegFormer~\cite{xie2021segformer}       & MiT-B1   & 0.536           & 0.533             & 13.7                    & 19.3                   \\
      SegFormer + WPRF                        & MiT-B1   & 0.590           & 0.605             & 13.7                    & 22.5                   \\
      \midrule
      Mask2Former~\cite{cheng2022mask2former} & Swin-T   & 0.568           & 0.580             & 29.8                    & 47.9                   \\
      Mask2Former + WPRF                      & Swin-T   & 0.600           & 0.621             & 29.9                    & 49.5                   \\
      \midrule
      TransUNet~\cite{chen2024transunet}      & resnet18 & 0.671           & 0.622             & 14.9                    & 42.2                   \\
      TransUNet + WPRF                        & resnet18 & 0.670           & 0.627             & 14.9                    & 43.0                   \\
      \midrule
      nnUNet~\cite{isensee2021nnunet}         & C64      & 0.805           & 0.637             & 15.9                    & 375.0                  \\
      nnUNet + WPRF                           & C64      & 0.802           & 0.638             & 15.9                    & 375.8                  \\
      \midrule
      CS-Net~\cite{mou2019csnet}              & C64      & 0.787           & 0.628             & 7.7                     & 185.7                  \\
      CS-Net + WPRF                           & C64      & 0.800           & 0.632             & 7.8                     & 186.5                  \\
      \midrule
      Swin-UNet~\cite{cao2023swinunet}        & Swin-T   & 0.801           & 0.670             & 28.0                    & 61.3                   \\
      Swin-UNet + WPRF                        & Swin-T   & 0.775           & 0.671             & 28.0                    & 62.1                   \\
      \midrule
      U-Mamba~\cite{ma2024umamba}             & C64      & 0.802           & 0.633             & 5.5                     & 136.0                  \\
      U-Mamba + WPRF                          & C64      & 0.816           & 0.635             & 5.5                     & 136.9                  \\
      \bottomrule
    \end{tabularx}
  }
\end{table}

\begin{table}[t]
  \centering
  \caption{Quantitative comparison on DeepCrack.}
  \label{tab:sota_deepcrack}
  {\setlength{\tabcolsep}{3pt}\scriptsize
    \begin{tabularx}{\linewidth}{l l *{4}{>{\raggedleft\arraybackslash}X}}
      \toprule
      Method                                  & Backbone & Dice $\uparrow$ & clDice $\uparrow$ & Params (M) $\downarrow$ & FLOPs (G) $\downarrow$ \\
      \midrule
      UNet~\cite{ronneberger2015u}            & resnet18 & 0.798           & 0.861             & 14.3                    & 34.4                   \\
      UNet + WPRF                             & resnet18 & 0.808           & 0.873             & 14.3                    & 34.6                   \\
      \midrule
      DeepLabv3+~\cite{chen2018deeplabv3plus} & resnet18 & 0.779           & 0.854             & 12.3                    & 18.5                   \\
      DeepLabv3+ + WPRF                       & resnet18 & 0.780           & 0.854             & 12.3                    & 19.5                   \\
      \midrule
      SegFormer~\cite{xie2021segformer}       & MiT-B1   & 0.777           & 0.847             & 13.7                    & 23.7                   \\
      SegFormer + WPRF                        & MiT-B1   & 0.782           & 0.850             & 13.7                    & 27.6                   \\
      \midrule
      Mask2Former~\cite{cheng2022mask2former} & Swin-T   & 0.778           & 0.852             & 29.8                    & 58.5                   \\
      Mask2Former + WPRF                      & Swin-T   & 0.808           & 0.882             & 29.9                    & 60.4                   \\
      \midrule
      TransUNet~\cite{chen2024transunet}      & resnet18 & 0.764           & 0.847             & 14.9                    & 51.0                   \\
      TransUNet + WPRF                        & resnet18 & 0.802           & 0.870             & 14.9                    & 52.0                   \\
      \midrule
      nnUNet~\cite{isensee2021nnunet}         & C64      & 0.790           & 0.852             & 15.9                    & 452.7                  \\
      nnUNet + WPRF                           & C64      & 0.798           & 0.861             & 15.9                    & 453.7                  \\
      \midrule
      CS-Net~\cite{mou2019csnet}              & C64      & 0.757           & 0.827             & 7.7                     & 224.1                  \\
      CS-Net + WPRF                           & C64      & 0.795           & 0.853             & 7.8                     & 225.1                  \\
      \midrule
      Swin-UNet~\cite{cao2023swinunet}        & Swin-T   & 0.777           & 0.867             & 28.0                    & 74.7                   \\
      Swin-UNet + WPRF                        & Swin-T   & 0.809           & 0.883             & 28.0                    & 75.7                   \\
      \midrule
      U-Mamba~\cite{ma2024umamba}             & C64      & 0.709           & 0.769             & 5.5                     & 164.2                  \\
      U-Mamba + WPRF                          & C64      & 0.746           & 0.787             & 5.5                     & 165.2                  \\
      \bottomrule
    \end{tabularx}
  }
\end{table}

\begin{table}[t]
  \centering
  \caption{Quantitative comparison on Massachusetts Roads.}
  \label{tab:sota_massroads}
  {\setlength{\tabcolsep}{3pt}\scriptsize
    \begin{tabularx}{\linewidth}{l l *{4}{>{\raggedleft\arraybackslash}X}}
      \toprule
      Method                                  & Backbone & Dice $\uparrow$ & clDice $\uparrow$ & Params (M) $\downarrow$ & FLOPs (G) $\downarrow$ \\
      \midrule
      UNet~\cite{ronneberger2015u}            & resnet18 & 0.729           & 0.839             & 14.3                    & 172.7                  \\
      UNet + WPRF                             & resnet18 & 0.733           & 0.843             & 14.3                    & 174.0                  \\
      \midrule
      DeepLabv3+~\cite{chen2018deeplabv3plus} & resnet18 & 0.702           & 0.810             & 12.3                    & 92.9                   \\
      DeepLabv3+ + WPRF                       & resnet18 & 0.708           & 0.815             & 12.3                    & 97.7                   \\
      \midrule
      SegFormer~\cite{xie2021segformer}       & MiT-B1   & 0.746           & 0.858             & 13.7                    & 172.5                  \\
      SegFormer + WPRF                        & MiT-B1   & 0.764           & 0.876             & 13.7                    & 191.8                  \\
      \midrule
      Mask2Former~\cite{cheng2022mask2former} & Swin-T   & 0.726           & 0.825             & 29.8                    & 290.6                  \\
      Mask2Former + WPRF                      & Swin-T   & 0.762           & 0.868             & 29.9                    & 300.2                  \\
      \midrule
      TransUNet~\cite{chen2024transunet}      & resnet18 & 0.728           & 0.841             & 14.9                    & 259.4                  \\
      TransUNet + WPRF                        & resnet18 & 0.731           & 0.843             & 14.9                    & 264.2                  \\
      \midrule
      nnUNet~\cite{isensee2021nnunet}         & C64      & 0.739           & 0.841             & 15.9                    & 2272.4                 \\
      nnUNet + WPRF                           & C64      & 0.751           & 0.855             & 15.9                    & 2277.2                 \\
      \midrule
      CS-Net~\cite{mou2019csnet}              & C64      & 0.739           & 0.844             & 7.7                     & 1124.9                 \\
      CS-Net + WPRF                           & C64      & 0.737           & 0.843             & 7.8                     & 1129.8                 \\
      \midrule
      Swin-UNet~\cite{cao2023swinunet}        & Swin-T   & 0.758           & 0.866             & 28.0                    & 372.3                  \\
      Swin-UNet + WPRF                        & Swin-T   & 0.755           & 0.862             & 28.0                    & 377.1                  \\
      \midrule
      U-Mamba~\cite{ma2024umamba}             & C64      & 0.702           & 0.805             & 5.5                     & 824.3                  \\
      U-Mamba + WPRF                          & C64      & 0.700           & 0.802             & 5.5                     & 829.2                  \\
      \bottomrule
    \end{tabularx}
  }
\end{table}

\begin{table}[t]
  \centering
  \caption{Quantitative comparison on OMVIS.}
  \label{tab:sota_omvis}
  {\setlength{\tabcolsep}{3pt}\scriptsize
    \begin{tabularx}{\linewidth}{l l *{4}{>{\raggedleft\arraybackslash}X}}
      \toprule
      Method                                  & Backbone & Dice $\uparrow$ & clDice $\uparrow$ & Params (M) $\downarrow$ & FLOPs (G) $\downarrow$ \\
      \midrule
      UNet~\cite{ronneberger2015u}            & resnet18 & 0.914           & 0.926             & 14.3                    & 258.1                  \\
      UNet + WPRF                             & resnet18 & 0.913           & 0.925             & 14.3                    & 259.9                  \\
      \midrule
      DeepLabv3+~\cite{chen2018deeplabv3plus} & resnet18 & 0.806           & 0.832             & 12.3                    & 138.8                  \\
      DeepLabv3+ + WPRF                       & resnet18 & 0.817           & 0.841             & 12.3                    & 146.0                  \\
      \midrule
      SegFormer~\cite{xie2021segformer}       & MiT-B1   & 0.803           & 0.833             & 13.7                    & 306.9                  \\
      SegFormer + WPRF                        & MiT-B1   & 0.844           & 0.870             & 13.7                    & 335.8                  \\
      \midrule
      Mask2Former~\cite{cheng2022mask2former} & Swin-T   & 0.880           & 0.889             & 29.8                    & 425.0                  \\
      Mask2Former + WPRF                      & Swin-T   & 0.881           & 0.890             & 29.9                    & 439.5                  \\
      \midrule
      TransUNet~\cite{chen2024transunet}      & resnet18 & 0.845           & 0.870             & 14.9                    & 390.8                  \\
      TransUNet + WPRF                        & resnet18 & 0.844           & 0.867             & 14.9                    & 398.0                  \\
      \midrule
      nnUNet~\cite{isensee2021nnunet}         & C64      & 0.808           & 0.791             & 15.9                    & 3395.3                 \\
      nnUNet + WPRF                           & C64      & 0.816           & 0.811             & 15.9                    & 3402.5                 \\
      \midrule
      CS-Net~\cite{mou2019csnet}              & C64      & 0.851           & 0.878             & 7.7                     & 1680.8                 \\
      CS-Net + WPRF                           & C64      & 0.868           & 0.879             & 7.8                     & 1688.0                 \\
      \midrule
      Swin-UNet~\cite{cao2023swinunet}        & Swin-T   & 0.876           & 0.887             & 28.0                    & 547.2                  \\
      Swin-UNet + WPRF                        & Swin-T   & 0.876           & 0.889             & 28.0                    & 554.4                  \\
      \midrule
      U-Mamba~\cite{ma2024umamba}             & C64      & 0.778           & 0.794             & 5.5                     & 1231.7                 \\
      U-Mamba + WPRF                          & C64      & 0.801           & 0.810             & 5.5                     & 1238.9                 \\
      \bottomrule
    \end{tabularx}
  }
\end{table}

\begin{figure}[t]
  \centering
  \includegraphics[width=\linewidth]{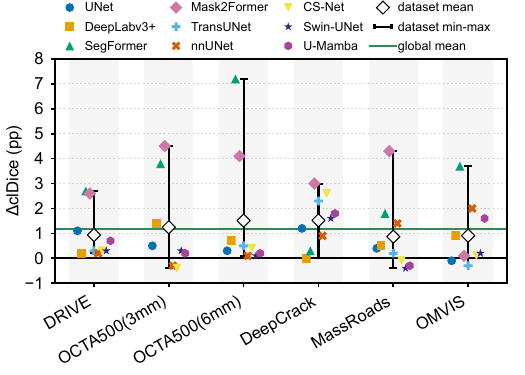}
  \caption{Aggregated main results from Tables~\ref{tab:sota_drive}--\ref{tab:sota_omvis}. Per-dataset distribution of $\Delta$clDice (pp) across nine backbones, where $\Delta$clDice $=$ clDice(+WPRF) $-$ clDice(baseline). Markers denote per-backbone values, whiskers span min--max, and diamonds indicate dataset means. The green line marks the global mean of +0.99\,pp (47/54 positive).}
  \label{fig:main_overview}
\end{figure}

Across six datasets and nine architectures, +WPRF improves clDice in 47 of 54 method--dataset pairs. The improvement magnitude correlates with structural fragility: on datasets with dense thin structures and severe connectivity bottlenecks such as OCTA-500 and DeepCrack, clDice gains reach +7.2\,pp (13.5\% relative improvement), while on datasets where baselines already achieve high connectivity, improvements remain modest yet stable. Max-Min gradient routing concentrates learning signals on the weakest connectivity links.

The gain magnitude correlates with structural fragility. The largest improvements appear on DeepCrack and OCTA-500 6\,mm, where structures are thin and connectivity bottlenecks are dense. WPRF's Max-Min gradient routing concentrates learning signals on these bottlenecks, repairing breaks without sacrificing region overlap.

Massachusetts Roads shows the smallest average improvement. Road structures are wider than vessels or cracks, and baseline connectivity is already high, leaving less room for bottleneck-focused repair. Even so, methods with weaker baselines still benefit, e.g., nnUNet gains +1.4\,pp in clDice.

\subsection{Qualitative Results}\label{sec:4-4}

\begin{figure*}[t]
  \centering
  \includegraphics[width=\linewidth]{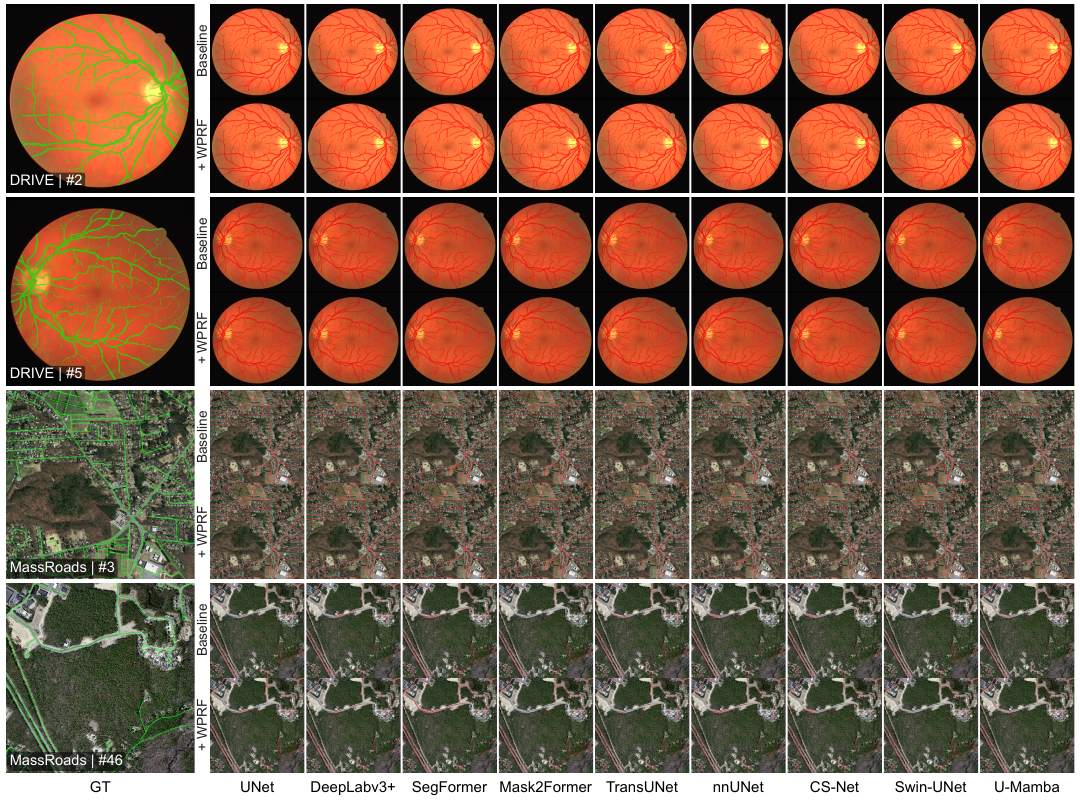}
  \caption{Qualitative comparison of nine architectures on DRIVE and Massachusetts Roads datasets. For each sample, top row: baseline prediction; bottom row: +WPRF prediction; GT overlaid in green. Baseline models exhibit topological breaks on thin structures, while WPRF consistently repairs these disconnections and preserves global connectivity.}
  \label{fig:qualitative}
\end{figure*}

\cref{fig:qualitative} shows qualitative comparison between baseline methods and their +WPRF variants across all nine architectures on DRIVE and Massachusetts Roads datasets.

Baseline models produce severe topological breaks on thin and low-contrast structures. For example, in DRIVE, small capillaries are often fragmented; in Massachusetts Roads, road networks are disconnected by occlusions or faint textures. By contrast, +WPRF variants consistently repair these disconnections and restore continuous topology of slender structures. Mechanistically, the $\min$ operator retains only the bottleneck-edge gradient along each path while the $\max$ operator selects the best path, so optimization focus shifts progressively from each repaired bottleneck to the next weakest link, forcing the network to prioritize repairing the weakest links during training and producing topologically complete predictions at inference.
This improvement is robust across different network paradigms, including CNNs, Transformers, and state-space models. Bottleneck-dominated gradient routing thus provides an architecture-agnostic mechanism for connectivity enhancement. These visual results corroborate the quantitative clDice improvements.

\subsection{Ablation Studies}\label{sec:4-5}
We use Mask2Former as the ablation base on DRIVE and Massachusetts Roads. Three loss baselines are included: BCE, Focal, and BCE+clDice. Starting from BA, we progressively add edge supervision, reachability supervision, and multi-scale reachability. \cref{tab:ablation} reports results.

\begin{table}[t]
  \centering
  \scriptsize
  \caption{Ablation study on DRIVE and Massachusetts Roads using Mask2Former (Swin-T). BA: bottleneck-aware segmentation term. Support: propagation domain for $\mathcal{L}_{\mathrm{reach}}$---full union occupancy $U^\ast_\Omega$ (unrestricted) or domain-restricted skeleton graph $V^\ast$. MS: multi-scale reachability ($\mathcal{K}=\{1,2,4,8,16\}$); single-scale otherwise ($\mathcal{K}=\{8\}$).}
  \label{tab:ablation}
  {\setlength{\tabcolsep}{3pt}
    \begin{tabularx}{\linewidth}{c c c c c X X X X}
      \toprule
      \multirow{2}{*}[-0.8ex]{$\mathcal{L}_{\mathrm{seg}}$} & \multirow{2}{*}[-0.8ex]{$\mathcal{L}_{\mathrm{edge}}$} & \multirow{2}{*}[-0.8ex]{$\mathcal{L}_{\mathrm{reach}}$} & \multirow{2}{*}[-0.8ex]{Support} & \multirow{2}{*}[-0.8ex]{MS} & \multicolumn{2}{c}{DRIVE} & \multicolumn{2}{c}{Mass.\ Roads}                  \\
      \cmidrule(l{0pt}r{2pt}){6-7}\cmidrule(l{2pt}r{0pt}){8-9}
                                                            &                                                        &                                                         &                                  &                             & Dice                      & clDice                           & Dice  & clDice \\
      \midrule
      BCE                                                   & --                                                     & --                                                      & --                               & --                          & 0.789                     & 0.805                            & 0.726 & 0.825  \\
      Focal                                                 & --                                                     & --                                                      & --                               & --                          & 0.786                     & 0.788                            & 0.721 & 0.822  \\
      BCE+clDice                                            & --                                                     & --                                                      & --                               & --                          & 0.794                     & 0.813                            & 0.728 & 0.829  \\
      BA                                                    & --                                                     & --                                                      & --                               & --                          & 0.795                     & 0.815                            & 0.727 & 0.829  \\
      BA                                                    & \checkmark                                             & --                                                      & --                               & --                          & 0.794                     & 0.815                            & 0.730 & 0.832  \\
      BA                                                    & \checkmark                                             & \checkmark                                              & $U^\ast_\Omega$                  & --                          & 0.796                     & 0.818                            & 0.730 & 0.845  \\
      BA                                                    & \checkmark                                             & \checkmark                                              & $V^\ast$                         & --                          & 0.797                     & 0.822                            & 0.742 & 0.845  \\
      BA                                                    & \checkmark                                             & \checkmark                                              & $V^\ast$                         & \checkmark                  & 0.796                     & 0.831                            & 0.762 & 0.868  \\
      \bottomrule
    \end{tabularx}
  }
\end{table}

Focal loss targets pixel-level hard examples but does not reliably improve connectivity. Pixel-level difficulty and topological criticality are distinct: Focal-weighted hard pixels are often blurry boundaries rather than connectivity-limiting bottlenecks.

BA achieves comparable clDice to the explicit skeleton loss (BCE+clDice) by up-weighting thin-structure pixels. Bottleneck-aware pixel reweighting thus complements skeleton-based supervision. This validates our core hypothesis: routing gradients to bottleneck pixels suffices to improve connectivity without explicit skeleton extraction.

Local edge supervision $\mathcal{L}_{\mathrm{edge}}$ yields stable gains on dense networks (Massachusetts Roads) but marginal improvement on sparse networks (DRIVE). Edge supervision thus primarily suppresses local breaks in dense regions. Reachability supervision $\mathcal{L}_{\mathrm{reach}}$ brings substantial gains on both datasets, confirming that long-range connectivity constraints are the core mechanism.

Restricting the propagation domain from the full occupancy graph $U^{\ast}_{\Omega}$ to the support graph $V^{\ast}$ further improves clDice on DRIVE and substantially boosts Dice on Massachusetts Roads. Domain restriction thus acts differently across morphologies: it suppresses spurious connections on sparse networks (improving clDice) and reduces background leakage on dense networks (improving Dice).

Multi-scale reachability improves clDice on both datasets, with larger gains on wider road networks (Massachusetts Roads: $+2.30$\,pp vs. DRIVE: $+0.90$\,pp). Multi-scale propagation is thus more critical for capturing long-range connectivity. The slight Dice drop on DRIVE suggests that stronger structural regularization trades a small amount of pixel overlap for better skeleton connectivity.

Different components contribute differently across morphologies: support-domain restriction is more critical for densely packed thin structures, while multi-scale reachability accounts for major gains on wider road networks. This complementarity enables WPRF to generalize across datasets.

\subsection{Analysis and Discussion}\label{sec:4-6}

We analyze WPRF from three angles: parameter sensitivity, gradient routing verification, and computational efficiency.

\subsubsection{Parameter Sensitivity}\label{sec:4-6-sens}

\begin{figure}[t]
  \centering
  \includegraphics[width=\linewidth]{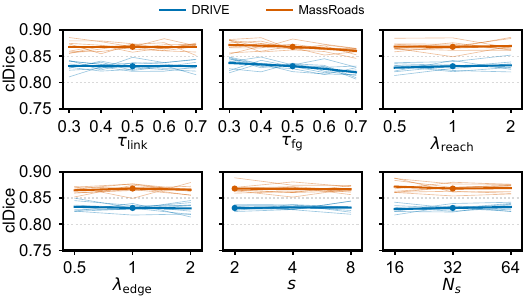}
  \caption{Hyperparameter sensitivity on DRIVE and Massachusetts Roads. Each subplot varies one parameter while holding all others at their default values. (a)~Link threshold $\tau_{\mathrm{link}}$, (b)~foreground threshold $\tau_{\mathrm{fg}}$, (c)~reachability loss weight $\lambda_{\mathrm{reach}}$, (d)~edge loss weight $\lambda_{\mathrm{edge}}$, (e)~graph construction stride $s$, (f)~number of sampled sources $N_s$.}
  \label{fig:sensitivity}
\end{figure}

We sweep six key hyperparameters on DRIVE and Massachusetts Roads to verify the robustness of WPRF near its default settings. Among them, $\tau_{\mathrm{link}}$ and $\tau_{\mathrm{fg}}$ are inference-time thresholds evaluated on a single trained model; the remaining four ($\lambda_{\mathrm{reach}}$, $\lambda_{\mathrm{edge}}$, $s$, $N_s$) require retraining. \cref{fig:sensitivity} shows the results.

\cref{fig:sensitivity} shows that clDice varies by less than 1\,pp across the tested ranges for all six hyperparameters. Performance is particularly stable within $\tau_{\mathrm{link}}\!\in\![0.3,0.7]$ and $\lambda_{\mathrm{reach}}\!\in\![0.5,2.0]$. For the graph stride, $s{=}4$ balances accuracy and efficiency: $s{=}2$ yields only marginal gain while quadrupling the node count, and $s{=}8$ loses fine-grained topological detail. The remaining parameters ($\tau_{\mathrm{fg}}$, $\lambda_{\mathrm{edge}}$, $N_s$) all produce near-flat curves, so fixed defaults suffice across datasets.

\subsubsection{Gradient Routing Mechanism}\label{sec:4-6-grad}

To verify the bottleneck-dominated gradient routing mechanism proposed in \cref{sec:1}, we diagnose per-pixel gradient distributions of baseline and +WPRF on a DRIVE training sample. Specifically, we record the gradient magnitude of the total loss with respect to the last decoder feature map, $g(x)=\lVert\partial\mathcal{L}/\partial\mathbf{f}(x)\rVert_2$, and average it over several epochs near the end of training to obtain $\bar{g}(x)$. To quantify gradient allocation across structures of varying thickness, we partition foreground pixels into five thickness bins via the distance transform of the ground-truth annotation and compute the fraction of total gradient energy in each bin.

\begin{figure}[t]
  \centering
  \includegraphics[width=\linewidth]{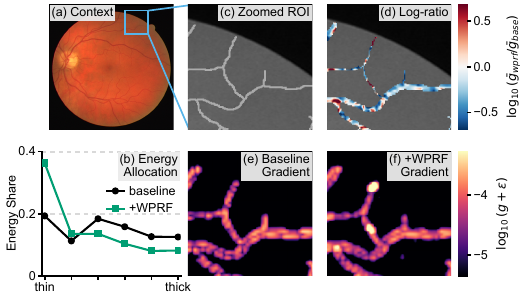}
  \caption{Gradient routing diagnosis on a DRIVE training sample. (a)~Full fundus image with region of interest (blue box). (b)~Gradient energy distribution across five vessel-thickness bins: +WPRF substantially shifts energy from thick to thin structures. (c)~Zoomed region of interest. (d)~Per-pixel log-ratio $\log_{10}(\bar{g}_{\mathrm{wprf}}/\bar{g}_{\mathrm{base}})$; red indicates stronger gradients under WPRF. (e), (f)~Gradient magnitude maps $\log_{10}(g+\epsilon)$ for baseline and +WPRF, respectively; +WPRF concentrates high-magnitude gradients at thin bottleneck junctions.}
  \label{fig:gradient_heatmap}
\end{figure}

\cref{fig:gradient_heatmap}(b) shows that baseline distributes gradient energy roughly evenly across thickness bins, with the thinnest bin receiving approximately 20\%. This lack of preferential attention to connectivity-critical bottlenecks is characteristic of TGS. With +WPRF, the thinnest bin's share rises from approximately 20\% to nearly 40\% while thicker bins decrease correspondingly, confirming that Max-Min algebra reroutes learning signals toward bottleneck pixels.

The spatial maps reinforce this picture. In \cref{fig:gradient_heatmap}(d), deep-red regions ($\log_{10}(\bar{g}_{\mathrm{wprf}}/\bar{g}_{\mathrm{base}})\!\approx\!{+}0.5$, i.e., WPRF gradients ${\approx}3\times$ stronger) coincide with thin side-branches and bottleneck junctions, while thick trunks appear blue or neutral. Panels~(e) vs.\ (f) show that baseline spreads gradients fairly uniformly along the skeleton, whereas +WPRF concentrates high-magnitude gradients at thin connecting bottlenecks. Mechanistically, the $\min$ operator retains only the bottleneck-edge gradient along each path while the $\max$ operator selects the best path, so optimization focus shifts progressively from each repaired bottleneck to the next weakest link---a seek-and-repair dynamic directly visualized by the hotspots in panel~(f).

\subsubsection{Computational Efficiency}\label{sec:4-6-eff}

WPRF adds overhead only at training time; no graph propagation runs at inference. \cref{tab:wprf_overhead} reports end-to-end overhead on DRIVE and Massachusetts Roads across three backbone families (CNN, Transformer, state-space).

\begin{table}[t]
  \centering
  \caption{Training and inference overhead of +WPRF on DRIVE and Massachusetts Roads. All measurements use batch size 1. Training time is reported per iteration; inference latency is reported per image. Peak memory denotes the maximum GPU memory allocated during training. All timings are averaged over 100 iterations after 20 warmup steps under CUDA synchronization, excluding data loading.}
  \label{tab:wprf_overhead}
  {\setlength{\tabcolsep}{6pt}\scriptsize
    \begin{tabularx}{\linewidth}{l l *{3}{>{\raggedleft\arraybackslash}X}}
      \toprule
      Method             & Backbone & Train (ms/iter)$\downarrow$ & Peak mem (GB)$\downarrow$ & Infer (ms/img)$\downarrow$ \\
      \midrule
      \multicolumn{5}{l}{\textit{DRIVE} (512$\times$512)}                                                                  \\
      \addlinespace[3pt]
      UNet               & resnet18 & 9.7                         & 0.9                       & 2.3                        \\
      UNet + WPRF        & resnet18 & 79.7                        & 1.0                       & 2.4                        \\
      \addlinespace[2pt]
      U-Mamba            & C64      & 34.4                        & 4.8                       & 11.8                       \\
      U-Mamba + WPRF     & C64      & 106.5                       & 4.9                       & 11.9                       \\
      \addlinespace[2pt]
      Mask2Former        & Swin-T   & 49.6                        & 2.1                       & 9.8                        \\
      Mask2Former + WPRF & Swin-T   & 122.3                       & 2.6                       & 10.0                       \\
      \midrule
      \multicolumn{5}{l}{\textit{Massachusetts Roads} (1024$\times$1024)}                                                  \\
      \addlinespace[2pt]
      UNet               & resnet18 & 25.9                        & 1.6                       & 6.6                        \\
      UNet + WPRF        & resnet18 & 228.8                       & 1.8                       & 6.7                        \\
      \addlinespace[2pt]
      U-Mamba            & C64      & 98.1                        & 8.8                       & 33.8                       \\
      U-Mamba + WPRF     & C64      & 301.4                       & 8.9                       & 34.1                       \\
      \addlinespace[2pt]
      Mask2Former        & Swin-T   & 141.7                       & 3.8                       & 27.9                       \\
      Mask2Former + WPRF & Swin-T   & 346.5                       & 4.7                       & 28.5                       \\
      \bottomrule
    \end{tabularx}
  }
\end{table}

Max-Min DP propagation increases per-iteration training time by roughly 2--3$\times$, yet peak memory grows by only 0.1--0.9\,GB. Inference latency is virtually unchanged because no graph operations run at test time. WPRF's connectivity gains thus incur no additional computational cost at deployment. The only additional parameters come from the lightweight affinity head (${\sim}$5--20K); the Params/FLOPs in the main tables exclude the training-time DP overhead.

\section{Conclusion}\label{sec:5}
In this paper, we identified TGS, the systematic under-supervision of connectivity-critical bottleneck pixels by pixel-wise losses, and proposed Widest-Path Reachability Fields (WPRF) to address it.
WPRF replaces sum-based gradient aggregation with Max-Min algebra, inducing a bottleneck-dominated gradient flow that concentrates learning on the weakest links limiting end-to-end connectivity.
Experiments across nine architectures and six datasets show that WPRF improves clDice in 47 out of 54 method--dataset pairs.
On structurally fragile scenarios, clDice gains reach 7.2 percentage points with 13.5\% relative improvement while preserving pixel-level overlap.
The strong correlation between gain magnitude and structural fragility validates the bottleneck-dominated gradient routing principle.
The $k$-step Max-Min propagation increases per-iteration training time by approximately 2--3$\times$, while inference latency remains nearly unchanged.
Our current formulation assumes tree-like or sparsely-looped structures; dense mesh topologies may require adaptive $k$-step strategies.
Future work will explore sparse Max-Min propagation to reduce training overhead and extend WPRF to 3D volumetric data and instance-level settings.

\bibliographystyle{IEEEtran}
\bibliography{references}

\end{document}